**RESEARCH**  Open Access

# Predicting customer's gender and age depending on mobile phone data

Ibrahim Mousa Al-Zuabi[*] , Assef Jafar and Kadan Aljoumaa

*Correspondence: ibrahim.alzuabi@hiast.edu.sy
Faculty of Information Technology, Higher Institute for Applied Sciences and Technology, Damascus, Syria

## Abstract

In the age of data driven solution, the customer demographic attributes, such as gender and age, play a core role that may enable companies to enhance the offers of their services and target the right customer in the right time and place. In the marketing campaign, the companies want to target the real user of the GSM (global system for mobile communications), not the line owner. Where sometimes they may not be the same. This work proposes a method that predicts users' gender and age based on their behavior, services and contract information. We used call detail records (CDRs), customer relationship management (CRM) and billing information as a data source to analyze telecom customer behavior, and applied different types of machine learning algorithms to provide marketing campaigns with more accurate information about customer demographic attributes. This model is built using reliable data set of 18,000 users provided by SyriaTel Telecom Company, for training and testing. The model applied by using big data technology and achieved 85.6% accuracy in terms of user gender prediction and 65.5% of user age prediction. The main contribution of this work is the improvement in the accuracy in terms of user gender prediction and user age prediction based on mobile phone data and end-to-end solution that approaches customer data from multiple aspects in the telecom domain.

**Keywords:**  Gender prediction, Age prediction, Customer behavior, Machine learning, Big data, Classification, CDR

## Introduction

Nowadays, the mobile phone is one of the fastest growing technologies in the developing world with global penetration rates reaching 90% [1]. This makes it a huge warehouse for customer's data. That is, every action taken by the customer (short message service (SMS), Call or Internet session) gets recorded within the telecom operator, in the so called (CDRs). There are many types of CDRs used mainly by telecom billing systems. CDR contains a lot of information, (type of event, who is involved in this event, datetime, cell identifier where this event has taken place). This raw data represents a valuable source for analyzing human and social behavior [2]. In the agricultural domain [3] mobile phone data is used to analyze mobility and seasonal activity patterns related to livelihood zones in Senegal, by creating mobility profiles for population and segmentation. While in energy domain [4] this data is used to analyze human activity, facilitate population growth estimation in rural areas and extrapolate electricity needs. In health sector [5, 6] mobile phone data is used to study the relation between human mobility and prevalence of a disease using mobile data. This data could be used to analyze human behavior and compute psychology-informed indicators to predict customer's personality [7]. Telecom industry is a fertile ground for many challenges that benefit mobile operators to improve their business and competition advantages in different domains. There are two major big data uses cases in the telecom domain:

**Network improvement**

Operators have always been concerned about network performance improvement. Resulting from using big data analytics, operators can identify troubles combined with root cause analysis [8], improve quality of experience (QoE) [9], perform real-time troubleshooting and fix network performance issues.

All kinds of self-organizing network (SON) automation such as provisioning, configuring and commissioning can be convenient to the traffic request and the changes in the environment based on the acumen gained from big data analytics [10]. It allows the operators prioritize the alarms which is very useful to save time and prevent service failure [11].

**Marketing, sales and customer loyalty**

Marketing and sales are considered the largest domain of big data usage in telecom industry. Suitable big data analytics allows the operators to create more intelligent marketing campaigns based on customer profiling and segmentation [12], and do sales analytics to improve the sales. It can also be utilized to get better results from marketing promotions, increase revenue and implement geo marketing and real-time marketing. For example, Globe Telecom (Telecommunications Company in the Philippines) uses big data analytics to improve effectiveness of promotions by 600% [13].

Attracting a new customer costs much more than keeping an existing one, so churn prediction and management have become a matter of great concern to mobile service providers. A mobile service provider wishing to retain their subscribers needs to be able to predict which of them may be at-risk of changing the operator [14].

In this research we propose a solution to solve a real problem in telecom operator. The problem being, telecom operators sometimes suffer from unreliable demographic data of their customers. This research introduces a solution, which employs different domains, like big data science, telecom, social strategies for gender and age prediction, as well as a comparison of machine learning methods and results. We worked with an end-to-end solution starting from data acquisition closed with web applications as user interface and with Infographic visualization such as maps and interactive querying data through query builder interface, including all related data processing such as extracting, loading and transforming (ELT) processes.

The rest of the paper is structured as follows: In "Related work" section, we present related works on user demographic prediction. "Methods" section describes the data set we used in this research as well as the feature extraction methods and Big Data life cycle. "Results and discussion" section, we describe our results on gender, age and evaluate our approach in real-world case study, also we present proposed framework. Finally, we conclude our work and describe future work in "Conclusion" section.

## Related work

Several works discussed gender and age identification in different domains with different methods, for example: in [15] twitter data is used to predict user demographic based on users' first name, or based on twitter pictures in [16], or based on text mining and analysis [17]. Whereas user demographics data can be predicted based on created features from browsing behavior [18]. However, the number of mobile users exceeds the number of social networks users, therefore many studies discussed predicting user demographics based on users' smartphone applications [19, 20]. Most of works that predict user demographic based on mobile phone data, also called CDR, relied on large training set to predict demographics. Felbo [2] has addressed advanced methodologies in machine learning and used deep learning algorithms to benefit from its efficiency in large dimensions features and let these algorithms do its job with feature engineering instead of hand-engineered features. However, large data set is used to train the model, that is the data set relied on 1,50,000 customers and contains more than 250 million records, with accuracy of 79% for identifying gender and 63.1% for age. Sarraute et al. [21] used data set containing 5,00,000 customers to extract behavioral and social network features of the user, and used principal component analysis (PCA) method to select the important features. The best accuracy achieved was 81.4% for predicting gender and 62.3% for predicting age. While Dong et al. [22] relied on social features of the customer (Degree Centrality, Triadic Closure, …) by studying call network and messaging network, and using double-dependent factor graph model to predict gender and age in a mobile phone social graph, and the results were: 0.8063 for predicting gender and 0.7132 for predicting age, using F1-Measure. Martinez et al. used small size of reliable data set, with 10,000 users, and by using multiple algorithms (SVM, Random Forest and K-means) the accuracy obtained was 80% when the percentage of predicted instances was reduced [23]. In [1] bandicoot[1] tool is used to extract more than 1400 behavioral features, with different categories, and tested those features with different algorithms such as random forest, SVM, KNN, and the accuracy of the model was 79.7% at best for predicting the gender at developing countries as in South Asia.

In this work we have focused on extracting suitable and dedicated features for Syrian society also, we extract features from multiple resources like customer services and contract.

## Methods

This section describes the data set used in this work as well as the feature extraction methods and big data life cycle.

### Data description and preparation

Problem understanding and data understanding phases helped us to determine the data sources, and define the important ones to extract relevant features. Data lake was developed as a single point for all data. However, 5 data sources were selected for our prediction models:

---
[1] http://bandicoot.mit.edu.

**Table 1 CDR sample fields**

| Call type | GSM (A) | GSM (B) | Direction | Cell identifier | Duration | Date | ... |
|---|---|---|---|---|---|---|---|
| Call | +963********8 | +963********5 | Out | C83 | 56s | 10/10/2018 23:30:26 | ... |
| Call | +963********5 | +963********8 | In | C203 | 56s | 10/10/2018 23:30:26 | ... |
| SMS | +963********9 | +963********3 | Out | C322 | Null | 10/10/2018 23:59:11 | ... |
| SMS | +963********3 | +963********9 | In | C164 | Null | 10/10/2018 23:59:11 | ... |
| ... | ... | ... | ... | ... | ... | ... | ... |

**Table 2 Sample of customer's services**

| GSM | Economy | Education | Health | Horoscopes | Technology | Sport | ... |
|---|---|---|---|---|---|---|---|
| +963********9 | 0 | 1 | 0 | 0 | 1 | 0 | ... |
| +963********5 | 0 | 0 | 1 | 1 | 0 | 0 | ... |
| +963********8 | 1 | 0 | 0 | 0 | 0 | 0 | ... |
| +963********3 | 0 | 0 | 0 | 0 | 0 | 0 | ... |
| ... | ... | ... | ... | ... | ... | ... | ... |

**Table 3 Sample of cells and sites database**

| Cell identifier | Site identifier | Longitude | Latitude | ... |
|---|---|---|---|---|
| C147 | S73 | **.******2 | **.******7 | ... |
| C23 | S119 | **.******0 | **.******6 | ... |
| C64 | S14 | **.******1 | **.******0 | ... |
| ... | ... | ... | ... | ... |

1. *CDRs* The CDRs data contain all actions that were taken by the customer with all their attributes. A data collector in the network switch captures the usage in the form of CDR. These raw CDRs are in turn converted by the mediation system into a format understandable by the Billing System. Table 1 shows sample of CDR fields:

2. *Customer's services* All services that the customer enrolled in have been collected and classified manually based on service type such as services related to political news, sports news and horoscopes, etc..., these categories are handled as customer features. As a result we got the table of customer's service. Table 2 is a sample.

3. *Customers' contract information* The information of customer contract has been fetched from CRM system, it contains the basic information about the customer (gender, age, location...) and customer subscriptions' information, as one customer may have more than one subscription (two or more GSMs) with different types of subscriptions: pre-paid, post-paid, 3G, 4G ... subscription.

4. *Cells and sites database* Telecom companies related data to sites, its components and operators are stored in relational database. This data is used to extract the spatial features. Table 3 illustrates the shape of anonymized data for this data source.

5. *Reliable dataset* Building such predictive system needs a sample, which contains the real demographics such as gender and age for each gsm in this sample, whatever the demographics of the gsm owner, because sometimes the real user and the gsm owner are not the same. This reliable sample is used for supervised learning algorithms' phases training and testing, however directed methods have been followed to collect data about more than 18,000 customers randomly, within about 6 months.

**Table 4 Age groups of the sample**

| Age group | Year range | Percentage (%) |
|---|---|---|
| A | 18–27 | 32 |
| B | 28–39 | 41 |
| C | 40–60 | 27 |

The sample data used to build the model contains 64% males and 36% females. The data has been divided according to the age into groups. Each age group represents different age stage. And after model training, the system will predict the age group of the customer. Table 4 shows the age group with the percentage density of this category within the sample:

**Feature extraction**

Working on feature engineering and extraction were done based on our search and guess. Therefore in memory-processing tools is used for processing and analytical purposes. However, 220 features extracted for each customer. These features are belonging to 6 features categories, each category is provided with examples (knowing that there are features which belong to more than one category).

- *Behavioral individual features* Average call duration per day, entropy of duration, standard deviation of outbound calls duration per day at daytime, average of duration of inbound calls per day at night, standard deviation of received SMS per day at worktime, etc. their number is about (200 features).
- *Social behavior features:* Number of contacts, entropy of contact, average transactions received per customer, standard deviation of sent transactions per contact, etc. (20 features).
- *Spatial and mobility features* Average of mobility, average of mobility in holidays, home area type, night antenna entropy, daytime antenna entropy, workday antenna entropy, holiday antenna entropy, etc. their number is about (21 features).
- *Temporal features* Per workdays (Sunday to Thursday) and holidays, or per daytime (9–16) and at night: average number of SMSs received per daytime (17–8) in holidays, etc. (165 features).
- *Types of services* enrolled in: tech news services, educational services, sport news services, political news services, entertainment services, etc., (13 features).
- *Contract information* tariff type, GSM type, (2 features).

The total number of the mentioned features is 421, but there are about 201 features belong to more than one category, so the total number of features is 220.

Feature extraction is a transformation of raw data into derived values suitable for modeling. All extracted features for gender and age model is belonging to one of two types of features, either statistical or categorical feature. The statistical functions we relied on to extract statistical features is:

- Probability: for example the feature "probability of SMS on holiday" calculated by the formula

$$probability = \frac{Total\ number\ of\ SMSs\ in\ holiday}{Total\ number\ of\ SMSs} \tag{1}$$

- Standard deviation: For example the feature "Standard deviation of call duration" calculated by the formula

$$s = \sqrt{\frac{1}{N-1} \sum_{i=1}^{N} (x_i - \bar{x})^2} \tag{2}$$

where N is the number of calls, $x_i$ is duration of call $i$ and $\bar{x}$ is the mean of calls duration.

- Percentage: For example the feature "percentage of customer transactions which is out of customer's area home" calculated by the formula

$$percent = \frac{number\ of\ transactions\ out\ of\ customer's\ area\ home}{Total\ number\ of\ transactions} * 100 \tag{3}$$

- Entropy is the sum of the probability of each feature value times the log probability of that feature value. For example the feature "entropy of duration" calculated by the formula

$$Entropy\ of\ duration = -\sum_{i=1}^{k} P_i . \log_2(P_i) \tag{4}$$

$$P_i = \frac{Total\ duration\ with\ contact\ i}{Total\ duration\ with\ all\ contacts} \tag{5}$$

Those are the statistical features also, categorical features are used:

- Categorical features: That can take on one of a limited, and fixed number of possible values, the number of this type of features is 15. For example the feature "has economic news service" which is take one value either 0 or 1.

The demographic features of contract data source such as gender and age were excluded, because they are not reliable and we need to predict customer's gender and age based on the behavior.

These features are the input for classification algorithms. however, the distributions of features among gender and age were studied at descriptive analysis phase, to gain more insight about customer's behavior as Figs. 1 and 2 which illustrate samples of statistical studies on those features.

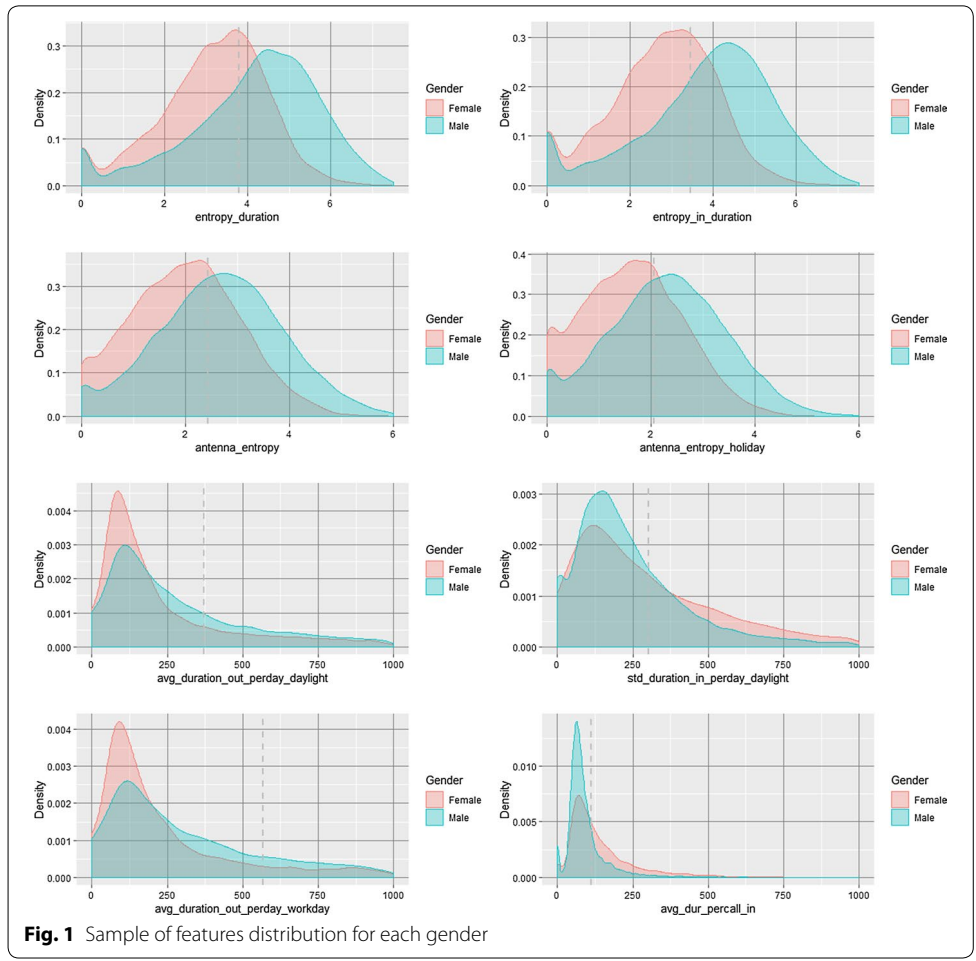
**Fig. 1** Sample of features distribution for each gender

As the statistical studies, some features values vary from male to female and from age group to another.

**Big data life cycle**

This work includes different stages (data exploration, features extraction and selection, and model validation), however different machine learning methodologies is used:

- *Unsupervised learning* For detecting outliers and dimensionality reduction like k-means clustering and PCA.
- *Supervised learning* Mainly classification methods for predicting customer's age and gender, therefore 12 of classification algorithms have been tested.

We have tested about 12 of the most popular classification algorithms. Regarding data acquisition and ELT process, we used multiple tools were used (Apache Flume,[2] Apache

---
[2] https://flume.apache.org.

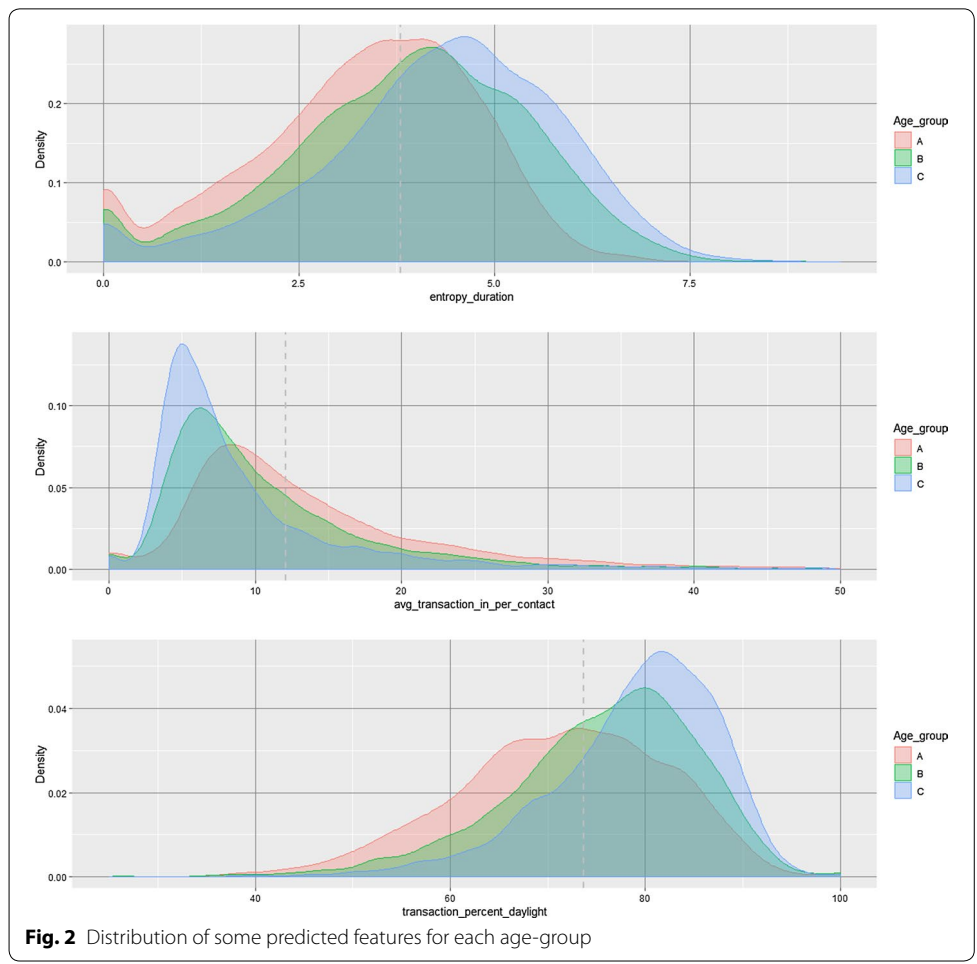

**Fig. 2** Distribution of some predicted features for each age-group

Nifi,[3] Apache sqoop[4]) to move the data, which have different sources and types (structured, semi-structured, streaming, and batches), from data sources and data lake to the big data platform. The data is stored within hadoop distributed file system (HDFS) [24] in the form of Apache Parquet storage format with Snappy compression for efficient storage and processing [25]. Extracting features from billions of CDRs' records was handled by Apache Spark [26], and the results were stored in Hadoop database HBASE[5], in order to be retrieved from the web applications.

The implemented framework represents the data life cycle phases and big data pipeline as shown in (Fig. 3).

---

[3] https://nifi.apache.org.

[4] http://sqoop.apache.org.

[5] https://hbase.apache.org

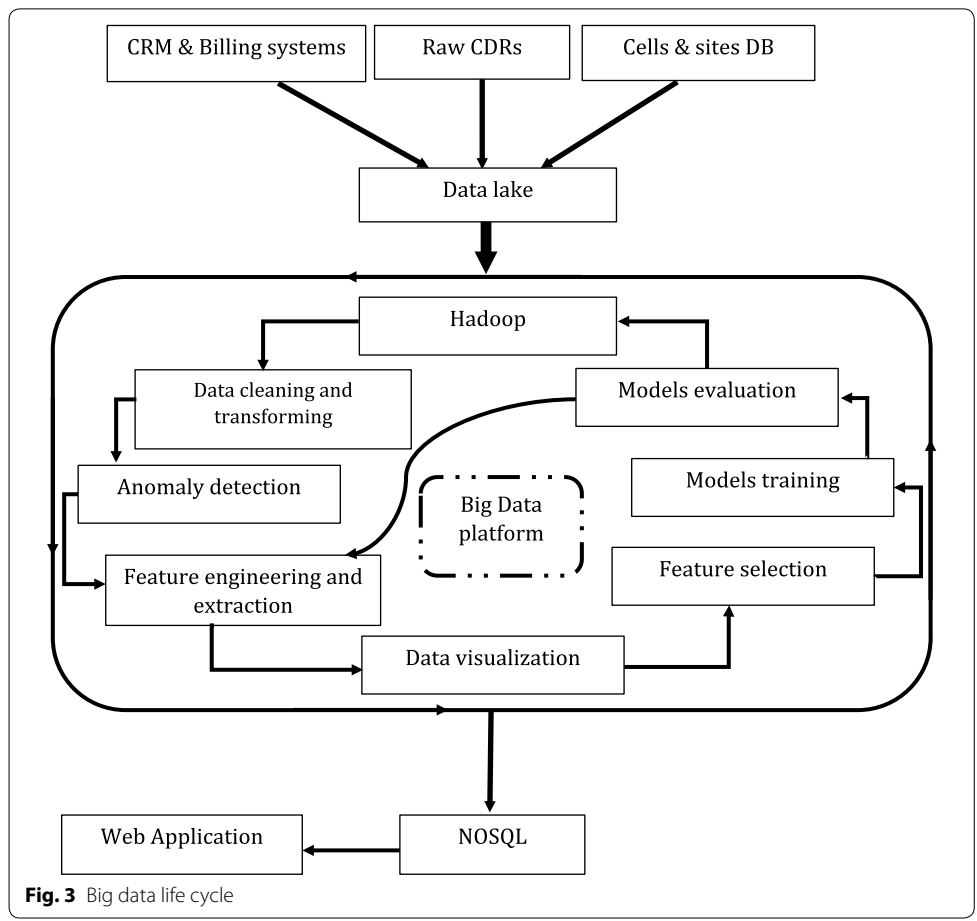

**Fig. 3** Big data life cycle

## Results and discussion

The dataset has been analyzed to recognizes customer types through the extracted features, customers with abnormal behavior have been detected using k-Means clustering, like customers having as average as more than 120 min of calls per day or having as average as more than 60 calls per day (later, we knew that their types of jobs explain their abnormality). Then, many classification algorithms were tested with extracted features. R language environment and its packages like caret and xgboost were used to preprocess those features and for modeling. The used preprocessing methods are:

- PCA: PCA with 10 and 100 principal components were tested, although it accelerated model's execution due to dimensionality reduction, but it didn't improve models results.

- Z-score, or standard score: Although it slightly improved SVM model, but it didn't improved the best model we got which is xgboost.

$$Z\text{-}score = \frac{x - \mu}{\sigma} \qquad (6)$$

The following classification algorithms were tested: linear discriminant analysis (LDA), support vector machine [(SVM (with a radial basis)], extreme gradient boosting (XGBoost), random forest, logistic regression, GLMNET, KNN, Naive Bayes, CART, C5.0, gradient boosting machine (GBM) and Bagged CART, however the best model is selected based on evaluation results.

For models training and validation, the reliable dataset was divided into (80%–20%). All classification algorithms have been trained using 10-fold cross validation, and relied on below metrics for model evaluation:

- *Accuracy* The number of correct predictions made is divided by the total number of predictions made. It is calculated by the formula

$$Accuracy = \frac{T_p + T_n}{P + N} \quad (7)$$

- *Area under the curve (AUC)* Measures classifier's performance [27]. It can be calculated by the formula

$$AUC = \int_0^1 TPR(x)dx \quad (8)$$

$$TPR = \frac{T_p}{T_p + F_n} \quad (9)$$

- *F1-measure* Harmonic mean of the precision and recall. It is calculated by the formula

$$F1\text{-}measure = \frac{2 * Precision * Recall}{(Precision + Recall)} \quad (10)$$

Regarding model's age evaluation, the formula of Mean F1 is

$$Mean\, F1 = \frac{F1\, grroup\,(A) + F1\, grroup\,(B) + F1\, grroup\,(C)}{3} \quad (11)$$

These metrics were used in this research to evaluate models on the testing set. Table 5 and Fig. 4 shows evaluation results regarding gender prediction. Table 6 and Fig. 5 shows best 4 evaluation results regarding age prediction, using big data platform (6 nodes, each node has processor of 16 cores and 32 GB of memory).

As a result, ensemble learning algorithms such as GBM, xgboost [28] and random forest, have more advantages on other classification algorithms and achieved best Accuracy, (AUC and F1-measure, that is xgboost score 0.8903 in F1-measure for gender prediction).

The tuning of xgboost on gender prediction model is (using xgboost package): max_depth = 10, eta = 0.1, gamma = 0, min_child_weight = 0.9, lambda = 0, alpha = 0.9, nrounds = 150, subsample = 1.

The tuning of xgboost for age prediction is: max_depth = 20, eta = 0.1, gamma = 0, min_child_weight = 0.9, lambda = 0, alpha = 0.9, nrounds = 50, subsample = 1.

**Table 5  Results for gender prediction**

| Algorithm | Accuracy | AUC | Classification time (s) |
| --- | --- | --- | --- |
| XGBoost | 0.8558 | 0.9226 | 0.3988 |
| Logistic regression | 0.8062 | 0.8727 | 0.3594 |
| Naive Bayes | 0.7084 | 0.7406 | 11.1174 |
| Random forest | 0.8396 | 0.9011 | 0.5021 |
| GBM | 0.8415 | 0.9039 | 0.9394 |
| Bagged CART | 0.8379 | 0.901 | 1.3061 |
| GLMNET | 0.7927 | 0.8669 | 0.2447 |
| KNN | 0.7233 | 0.7641 | 9.9087 |
| C5.0 | 0.8317 | 0.9023 | 1.4531 |
| CART | 0.7295 | 0.7627 | 0.4628 |
| LDA | 0.7813 | 0.8543 | 12.6713 |
| SVM | 0.8149 | 0.8796 | 1.5702 |

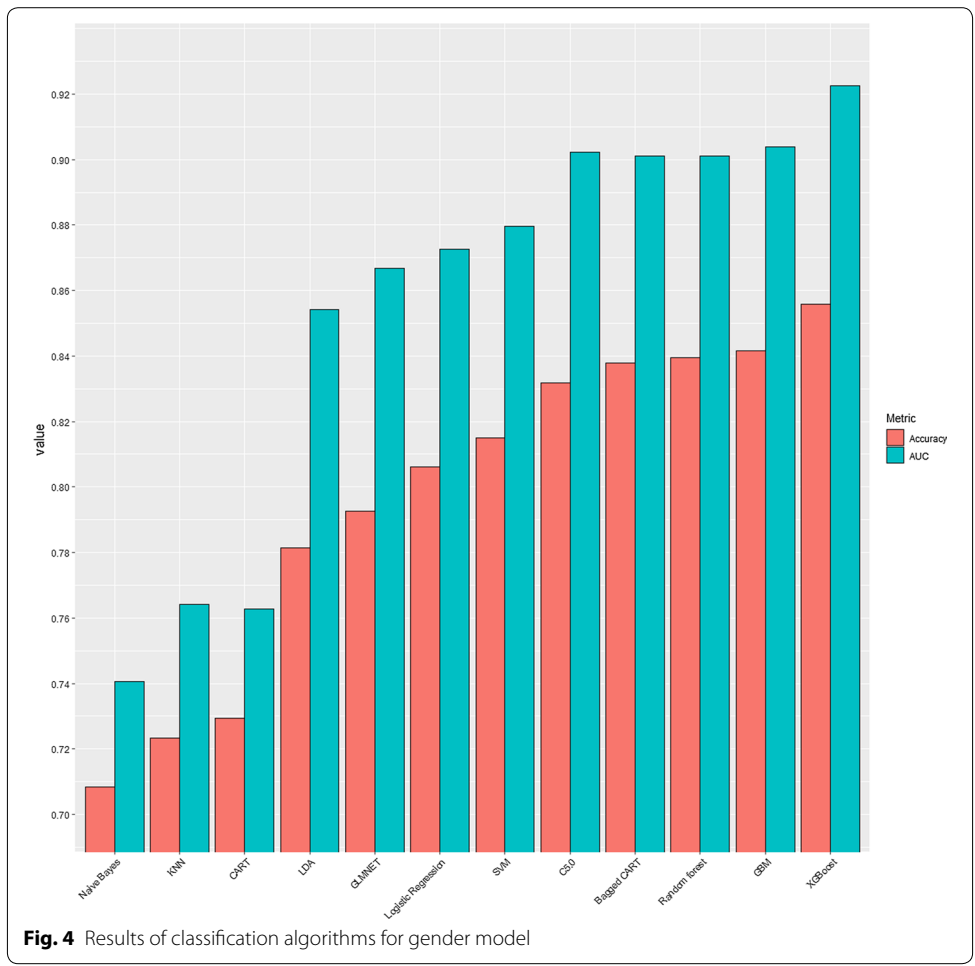

**Fig. 4** Results of classification algorithms for gender model

Increasing the number of trees more than 150 trees for xgboost in gender prediction didn't improve gender model accuracy. Also increasing the number of trees more than 50 didn't improve age model accuracy. That is at model learning process, each time we added

### Table 6 Results for age prediction

| Algorithm | Accuracy | Mean F1 | Classification time (s) |
|---|---|---|---|
| XGBoost | 0.655 | 0.6512 | 0.4596 |
| Random forest | 0.6428 | 0.6380 | 0.4920 |
| GBM | 0.626 | 0.6196 | 1.2764 |
| Bagged CART | 0.6331 | 0.6301 | 1.5232 |

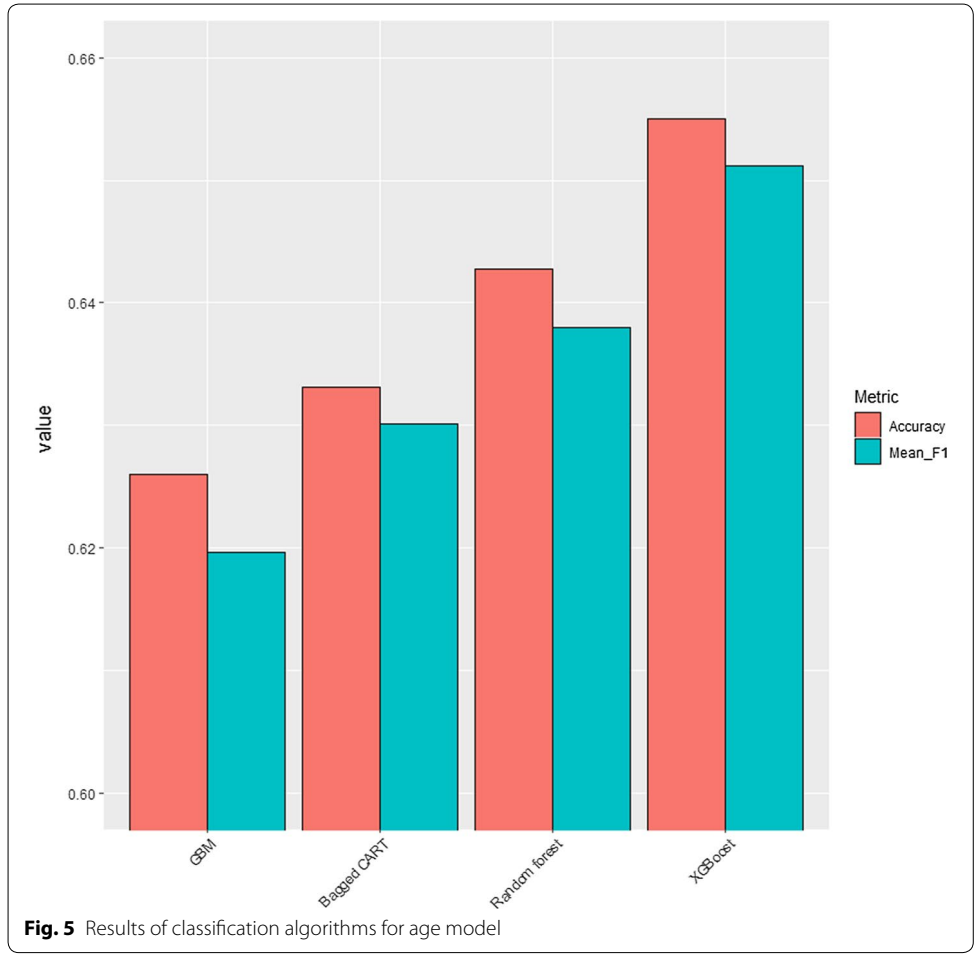

**Fig. 5** Results of classification algorithms for age model

tree to the xgboost model, the error rate is being tested on training and testing set, if the error rate on test set doesn't decrease, learning process should be stopped, even if the error rate of the training set continued to decline, because most likely the model is going to overfit.

Gain metric is considered as a measure for features importance, therefore it used to detect informative features for gender and age models.

*Gain* implies the relative contribution of the corresponding feature to the model calculated by taking each feature's contribution for each tree in the model. A higher value of this metric when compared to another feature implies it is more important for generating a prediction. It is calculated by the formula [28]

$$Gain = \frac{1}{2}\left[\frac{\left(\sum_{i\in I_L} g_i\right)^2}{\sum_{i\in I_L} h_i + \lambda} + \frac{\left(\sum_{i\in I_R} g_i\right)^2}{\sum_{i\in I_R} h_i + \lambda} - \frac{\left(\sum_{i\in I} g_i\right)^2}{\sum_{i\in I} h_i + \lambda}\right] - \gamma \quad_{I = I_L \cup I_R} \quad (12)$$

$$g_i = \frac{\vartheta L(Y, f(x))}{\vartheta f(x)}\bigg|_{f(x)=f^{(m-1)}(x)} \quad (13)$$

$$h_i = \frac{\vartheta^2 L(Y, f(x))}{\vartheta f(x)^2}\bigg|_{f(x)=f^{(m-1)}(x)} \quad (14)$$

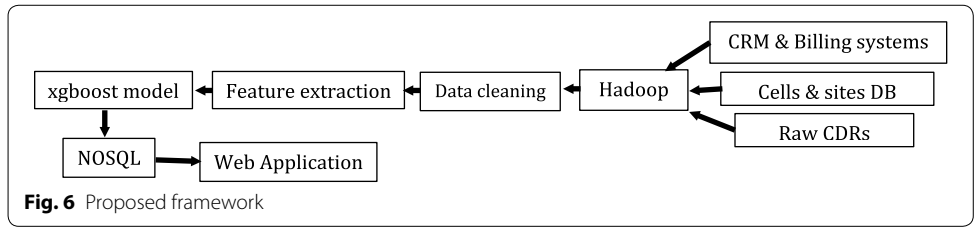

**Fig. 6** Proposed framework

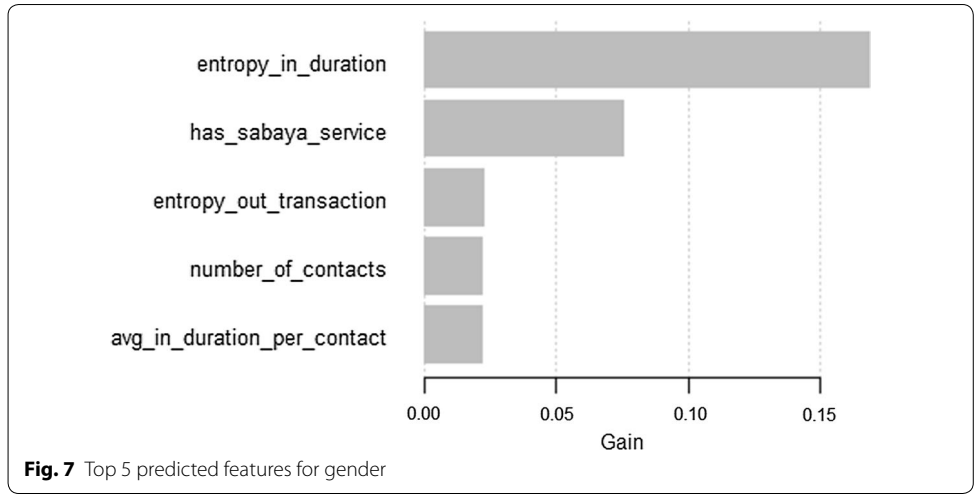

**Fig. 7** Top 5 predicted features for gender

The proposed framework was selected based on the comparison between the results of classification methods as Fig. 6.

According to xgboost model, the top predicted features for gender model based on gain measure are entropy of duration for received calls, services that were oriented for young girls. Figure 7 shows gain measure for top 5 informative features for gender prediction. Figure 8 shows gain measure for top 4 informative features for age prediction, like entropy of all calls' durations, average SMSs sent by the customer, etc…

These results reflect the nature of our conservative society where males usually bear responsibilities more than females, so that the males handle several types of contacts (business, family, friends, …). This justifies entropy in their telecommunication behavior (Figs. 1, 5). Also; with age model in (Figs. 2, 7) shows that group (A) which contains young people at the university age have less entropy because of having less contact types. The older the age, the more entropy we find in their telecommunication behavior compared to group (B) and (C). Less average transactions per contact can be justified by the increase of commitments towards more people and bigger families, which is a common thing in our country.

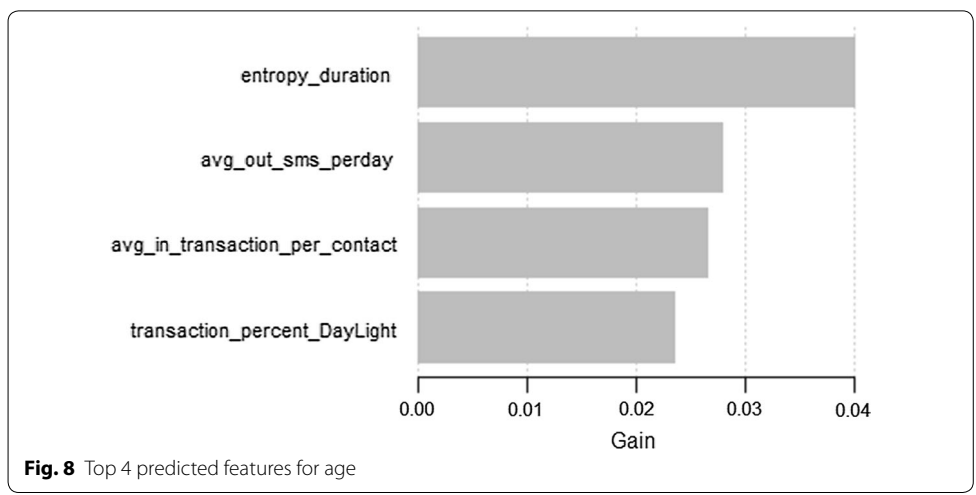

**Fig. 8** Top 4 predicted features for age

The limitation of this work is collecting reliable data (for training and testing) from random customers (only about 18000 customers) took a lot of time (about 6 months) because of following the direct methods and limitation of human resources for this process.

This work could be improved by being extended to include 2 new age-related groups which aren't included right now, one for people who are less than 18 years old and another one for people who are above 60 years old, and achieve a more balanced percentage regarding the gender to be more equal.

In addition, this work was conducted on two types of CDRs only, (calls CDR and SMS CDR), that we couldn't handle other types of CDR due to storage and process limitations. Internet usage CDR is considered as another data source to extract more valuable features, if the work gets rid of previously mentioned limitations, the reliable data set would be larger and more suitable for deep learning algorithms and the models will be more robust and accurate.

Another limitation is that, this work has been applied in the Syrian society, which may differ from other societies, so the informative features in this study could be more or less important in other societies.

## Conclusion

This work needed a lot of effort to analyze telecom customer behavior, based on their actions, services and contract information. The main contribution of this work is the end-to-end solution that approaching customer data from multiple aspects in the telecoms domain. Starting from creating enormous number of features combined with different types (behavioral, social, spatial, mobility, temporal, services and contract features). This grants us the opportunity to have deep insight on the customers. Then the relation between the age and gender with variety of informative attributes were studied based on statistical properties.

Another research direction is to explore more attribute related to the customer, and use these results to create dynamic offers system with an intelligence high enough to recommend and customize personalized offers.

#### Abbreviations
GSM: global system for mobile communications; CDR: call detail record; CRM: customer relationship management; SMS: short message service; QoE: quality of experience; SON: self-organizing network; ELT: extract, load and transform; PCA: principal component analysis; SVM: support vector machine; KNN: k-nearest neighbors; HDFS: Hadoop Distributed File System; LDA: linear discriminant analysis; XGBoost: extreme gradient boosting; GLMNET: lasso and elastic-net regularized generalized linear models; CART: classification and regression trees; GBM: gradient boosting machine; Bagged CART: bagging classification and regression trees; AUC: area under the curve; avg: average; std: standard deviation; in: received by the customer; out: send by the customer; $T_p$: *true positive*; $T_n$: *true negative*; $F_n$: *falsenegative*; $P$: *true positive + false negative*; $N$: *truenegative + falsepositive*; TPR: *truepositiverate*; $\mu$: mean of the population; $\sigma$: standard deviation of the population..

#### Authors' contributions
IMA-Z took on the main role so he performed the literature review, implemented the proposed model, conducted the experiments and wrote the manuscript. AJ and KJ took on a supervisory role and oversaw the completion of the work. All authors read and approved the final manuscript.

#### Acknowlegdements
This research was sponsored by SyriaTel telecom Co. We thank our colleagues Mem. Mjida (SyriaTel CEO), Mr. Murid (MIS director), Mr. Adham (Big Data manager) who provided insight and expertise that greatly assisted the research, although they may not agree with all of the interpretations/conclusions of this paper. Also, the authors thank Mustafa Mustafa, Hazem Deeb, MhdBahar Dasouki, Omran Abbas, Rawad Melhem, Maha Alkhayrat, Leen Taha, Mahmoud Lila, Bashar Wannows, Mhd Assaf, SyriaTel D3M unit for great ideas, help with the data processing and their useful discussions...

#### Competing interests
The authors declare that they have no competing interests.

#### Availability of data and materials
The data that support the findings of this study are available from SyriaTel Telecom Company but restrictions apply to the availability of these data, which were used under license for the current study, and so are not publicly available. Data are however available from the authors upon reasonable request and with permission of SyriaTel Telecom Company.

#### Consent for publication
The authors consent for publication.

#### Ethics approval and consent to participate
The authors Ethics approval and consent to participate.

#### Funding
The authors declare that they have no funding.

#### Publisher's Note
Nature remains neutral with regard to jurisdictional claims in published maps and institutional affiliations.

Received: 31 October 2018   Accepted: 28 January 2019

Published online: 19 February 2019